\documentclass[letterpaper, 10 pt, conference]{ieeeconf}  

\IEEEoverridecommandlockouts                              

\usepackage{graphicx} 
\usepackage{amsmath} 
\usepackage{amssymb}  
\usepackage[style=numeric,backend=bibtex]{biblatex} 
\usepackage{siunitx}  
\usepackage[nolist]{acronym} 
\usepackage[acronym,nonumberlist,nopostdot,nogroupskip]{glossaries}
\usepackage{float} 
\usepackage{caption}

\bibliography{references_no_url.bib}
\bibliography{references_other.bib}
\bibliography{bibliography_extra.bib}

\newtoggle{anonimo}
\togglefalse{anonimo}

\title{\LARGE \bf
How to rewrite the stars:\\ Mapping your orchard over time through constellations of fruits
}

\iftoggle{anonimo}{
}{
\author{Gonçalo P. Matos$^{1,2}$, Carlos Santiago$^{2}$, João P. Costeira$^{2}$, Ricardo L. Saldanha$^{1}$, Ernesto M. Morgado$^{1}$
\thanks{$^{1}$SISCOG -- Sistemas Cognitivos, SA}%
\thanks{$^{2}$Institute for Systems and Robotics (ISR) / LARSyS, Instituto Superior Técnico (IST), Lisbon, Portugal
        {\tt\small goncalo.p.matos@tecnico.ulisboa.pt}}%
}
}

\begin{document}

\thispagestyle{empty}
\pagestyle{empty}

\twocolumn[{%
\renewcommand\twocolumn[1][]{#1}%
\maketitle
\begin{center}
    \centering
    \captionsetup{type=figure}
    \includegraphics[width=1.0\textwidth]{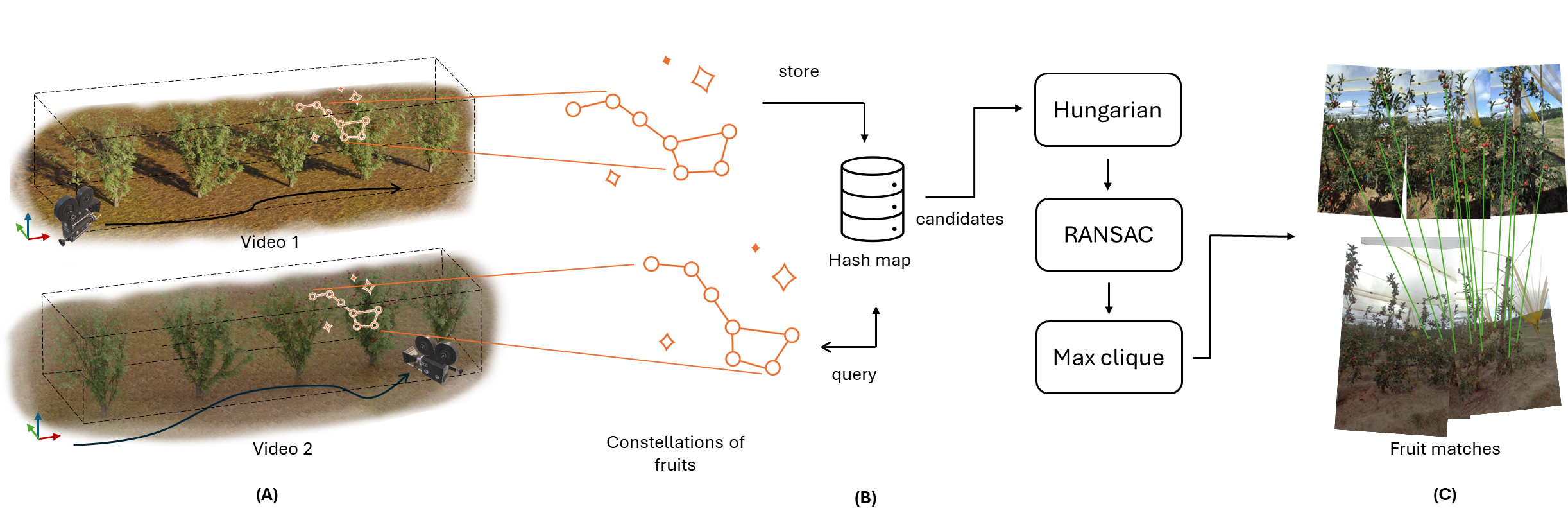}
    \captionof{figure}{Overview of the proposed method. (A) Fruits are tracked throughout the image sequence in stereo 3D. (B) Constellations of 3D fruit centroids are matched using the proposed descriptor. (C) Fruits are re-identified across videos.}
    \label{fig:teaser}
\end{center}%
}]

\begin{abstract}

Following crop growth through the vegetative cycle allows farmers to predict fruit setting and yield in early stages, but it is a laborious and non-scalable task if performed by a human who has to manually measure fruit sizes with a caliper or dendrometers. 
In recent years, computer vision has been used to automate several tasks in precision agriculture, such as detecting and counting fruits, and estimating their size. 
However, the fundamental problem of matching the exact same fruits from one video, collected on a given date, to the fruits visible in another video, collected on a later date, which is needed to track fruits' growth through time, remains to be solved. Few attempts were made, but they either assume that the camera always starts from the same known position and that there are sufficiently distinct features to match, or they used other sources of data like GPS.
Here we propose a new paradigm to tackle this problem, based on constellations of 3D centroids, and introduce a descriptor for very sparse 3D point clouds that can be used to match fruits across videos. Matching constellations instead of individual fruits is key to deal with non-rigidity, occlusions and challenging imagery with few distinct visual features to track.
The results show that the proposed method can be successfully used to match fruits across videos and through time, and also to build an orchard map and later use it to locate the camera pose in 6DoF, thus providing a method for autonomous navigation of robots in the orchard and for selective fruit picking, for example.

\end{abstract}



\section{INTRODUCTION}

In the context of precision agriculture, measuring fruits' growth over time provides insightful information to farmers, helping to predict what will be the fruit yield and to adjust agricultural practices, such as fruit thinning, to optimise the quality and quantity of production. Outliers in growth rates could also be symptoms of disease or irrigation anomalies. 
However, manually measuring and tracking fruits' sizes with calipers or dendrometers is a laborious process that is not scalable for large orchards, and sampling them to extrapolate results is error-prone. 

In recent years, computer vision techniques have been applied to precision agriculture, automating tasks such as fruit counting \cite{stein_image_2016,santos_pipeline_2023,matos_tracking_2024_anonymized,santos_apple_2025_anonymized} for yield estimation and fruit sizing \cite{checola_apple_2025}.
These applications generally address the problem as an isolated act of analysing the fruit present in one video shot on a given date, but tracking the crop's growth and evolution at fruit level through time requires capturing and relating multiple videos on different dates.
However, matching individual fruits across videos is a challenging task, since their visual appearance and their surroundings (trees' branches and leaves) change over time and in non-rigid ways, making it hard for common visual feature descriptors such as \ac{SIFT} \cite{lowe_distinctive_2004} to perform reliably. To our knowledge, this problem has not yet been solved in general settings, without assuming something about the initial pose and trajectory of the camera, or without the aid of external sensors like \ac{GPS} and \acp{IMU}.

The difficulty of finding distinct visual features to match images in cluttered and dynamic scenarios also hinders the task of creating a map for robots to navigate visually through the orchard, which could be useful to automate the selective application of chemicals or to perform fruit picking on only some trees.

In this work, we propose a different paradigm, which we call the \textit{constellation paradigm}, inspired by previous works with 2D semantic maps \cite{toso_you_2023,toso_maps_2024}. Instead of trying to find distinct features and matching them individually across images, we use semantic information --- the 3D location of fruits, obtained with an object detector and a stereo camera --- to lift the problem to a semantic sparse point cloud of object centroids, and then find a correspondence between small patterns of points, which we call \textit{constellations}, similarly to how people identify constellations of stars in the night sky. 

The robustness of the method comes from (1) using semantic information that is less sensitive to changes in visual appearance between videos, and (2) recognising distinctive patterns (constellations) in the relative positions of a few static objects, which produces less false matches than trying to match individual points.

We also contribute a new descriptor for very sparse 3D point clouds that is invariant to rotation, translation, and scale, which we use to match constellations of fruits across two 3D scenes with unknown relative pose and reconstruction scale.

\section{RELATED WORK}

The problem of finding visual correspondences between scenes that change appearance through time, due to illumination or seasonal changes, 
can be related to other broader problems known in the field of robotics as \textit{hierarchical localisation} \cite{sarlin_coarse_2019}, \textit{long-term visual localisation} \cite{toft_long-term_2022}, and \textit{place recognition} \cite{cieslewski_point_2016}. 
In many of these problems, a urban scene or an indoor scene, composed of mostly static and rigid objects with clear surfaces and edges, is usually the addressed scenario. 
Yet, in our case, we are dealing with objects (plants) that have no clear boundaries, surfaces or edges, which grow and change appearance in a non-rigid way over time.

Hence, trying to match visual features or dense point clouds in these settings is not very effective, which is why we lift the problem to the semantic level of constellations of fruits and try to match these very sparse semantic point clouds. 
Therefore, we relate our approach with other works that specifically addressed crop monitoring through time, or to approaches that involve matching sparse point clouds.

\subsection{Crop monitoring through time}

Lei et al. \cite{lei_4d_2024} proposed a method to monitor orchards through time at individual tree and fruit level in order to plot fruit growth charts, but their method fuses data from an RGB camera, a \ac{LiDAR} and an \ac{IMU} in order to build a 3D semantic map. Several of these semantic maps are then registered with the \ac{ICP} algorithm, which requires a good initial pose estimation. The authors work-around the initialisation problem by placing a physical marker on the ground and always starting the data collection procedures with the camera/sensors on that same place, which is not a viable solution to be used ``in the wild'' by farmers with cameras mounted on agricultural vehicles.

Dong et al. \cite{dong_4d_2017} also addressed the problem of monitoring the same 3D plant structures over time. 
They use a multi-sensor \ac{SLAM} pipeline, combining RGB imagery, \ac{IMU} and \ac{GPS} in a factor graph, to achieve a 3D reconstruction of each row of trees, but their approach strongly relies on the \ac{GPS} and \ac{IMU} measurements to first roughly register two reconstructions and to later guide the search for image correspondences between them with \ac{SIFT} \cite{lowe_distinctive_2004}. Moreover, the imagery that can be in seen in \cite{dong_4d_2017} suggests a considerable distance from the camera to the peanut plants, showing large areas of terrain and several rows of plants both in the foreground and background from which it is possible to extract distinctive features to match. In our case, though, we have a camera mounted in a tractor, facing a row of trees just about 1 meter away.

Chebrolou et al. \cite{chebrolu_robust_2018} focused on the problem of long-term registration of point clouds for crop monitoring using \ac{UAV} imagery. They propose a 2D geometric descriptor that outperforms \ac{SIFT} in the task of matching features across images captured at different dates, and use those matches in a \ac{BA} problem to jointly compute the camera poses and 3D positions of landmarks in two different dates.
However, their method requires a top-down and far view of the crop, where several plants are visible and easily discernible, and also assumes that the terrain is approximately planar.

\subsection{Matching of (sparse) point clouds}

In recent years, several descriptors were proposed for point clouds and 3D points in the context of point cloud registration \cite{borges_local_2020}, place recognition \cite{cieslewski_point_2016} or localisation \cite{he_m2dp_2016}.
However, they usually refer to dense point clouds obtained by \ac{LiDAR} sensors mounted on vehicles, which depict mostly static and rigid scenes, such as cities, and from which surfaces and normals can be estimated \cite{sun_efficient_2020,zhao_hoppf_2020}.

In contrast, our point clouds of trees obtained from stereo, besides being noisier than those obtained with \ac{LiDAR}, are cluttered, do not contain any planar surface, making it hard to compute normals, and hardly contain any distinctive features to track or match. 

Tinchev et al. \cite{tinchev_seeing_2018} propose a descriptor specifically for cluttered vegetation scenes, but they use very dense and precise \ac{LiDAR} point clouds, and they are not interested in identifying very small features at fruit level, but only in performing global localisation. Moreover, the authors note that the method struggled in a scene with uninterrupted bushes, and that it may fail if the sensor vantage point changes significantly between captures.

If we consider our very sparse semantic point clouds resulting from fruit centroids instead, we cannot apply most descriptors designed for dense point clouds. 
Even descriptors that were specifically proposed for sparse point clouds, such as PGHCI \cite{xu_heterogeneous_2022}, NBDL \cite{cieslewski_point_2016} or GLAROT-3D \cite{rizzini_place_2017}, usually work by dividing points in bins and computing relative densities or histograms, which still require some hundreds of points to extract a meaningful descriptor. As we shall see, in our case we have very few landmarks, with semantic point clouds that can be as low as 5 points, for which even these descriptors would not be effective.

Some of the limitations commonly identified in the literature for place recognition methods include the lack of robustness of some descriptors to point of view changes, changes in appearance caused by varying light conditions, weather, or season, and occlusions in the point cloud caused by other objects, which are all issues we face in our data. These motivated several authors \cite{dube_segmatch_2017,fan_seed_2020,zhu_gosmatch_2020,li_sa-loam_2021} to use semantic point clouds or graphs instead to tackle the place recognition and loop closure problem in \ac{SLAM} in cluttered and dynamic scenes, which also seems a promisor approach to deal with noisy orchard point clouds like ours. This motivated our lifting of the problem to the semantic level.

\section{PROPOSED METHOD}

\subsection{Overview}

The foundations for the ``constellation paradigm'' introduced before are to consider sets of fruits to form ``constellations'', i.e. small point clouds, and to develop a descriptor for such ``constellations'', in order to match them. 

Assuming that we have a method to obtain 3D representations of fruits as points corresponding to their centroids --- in our case, we use an implementation of the 3D tracking algorithm described in \cite{matos_tracking_2024_anonymized}  ---, building a ``constellation'' of fruits amounts to forming groups of a fixed number of $k$ such points and then encoding them with a descriptor. To group these points, we traverse the whole point cloud and consider one point (fruit) at a time, and find its $k$ nearest neighbours.
This set of $k$ points becomes one constellation, and is encoded into a descriptor vector.

Fruits are then matched across videos by finding a constellation with similar descriptor. Our descriptor allows the computation of a rigid transformation between the two matched constellations, hence it is possible to compute a relative camera pose, and localise a robot with respect to a map, for example.  

\subsection{STaR-i -- A descriptor for sparse 3D constellations}

STaR-i ---\textit{ Scale, Translation and Rotation-invariant} descriptor for very sparse 3D point clouds --- is inspired by an astrometry solver published by Lang et al. \cite{lang_astrometrynet_2010}.
Their descriptor works for constellations of stars in 2D images of the night sky and is invariant to translations, rotations, and scale. Our descriptor extends it to 3D while preserving these same properties.

\begin{figure*}[ht]
  \centerline{
  \includegraphics[width=0.3\linewidth]{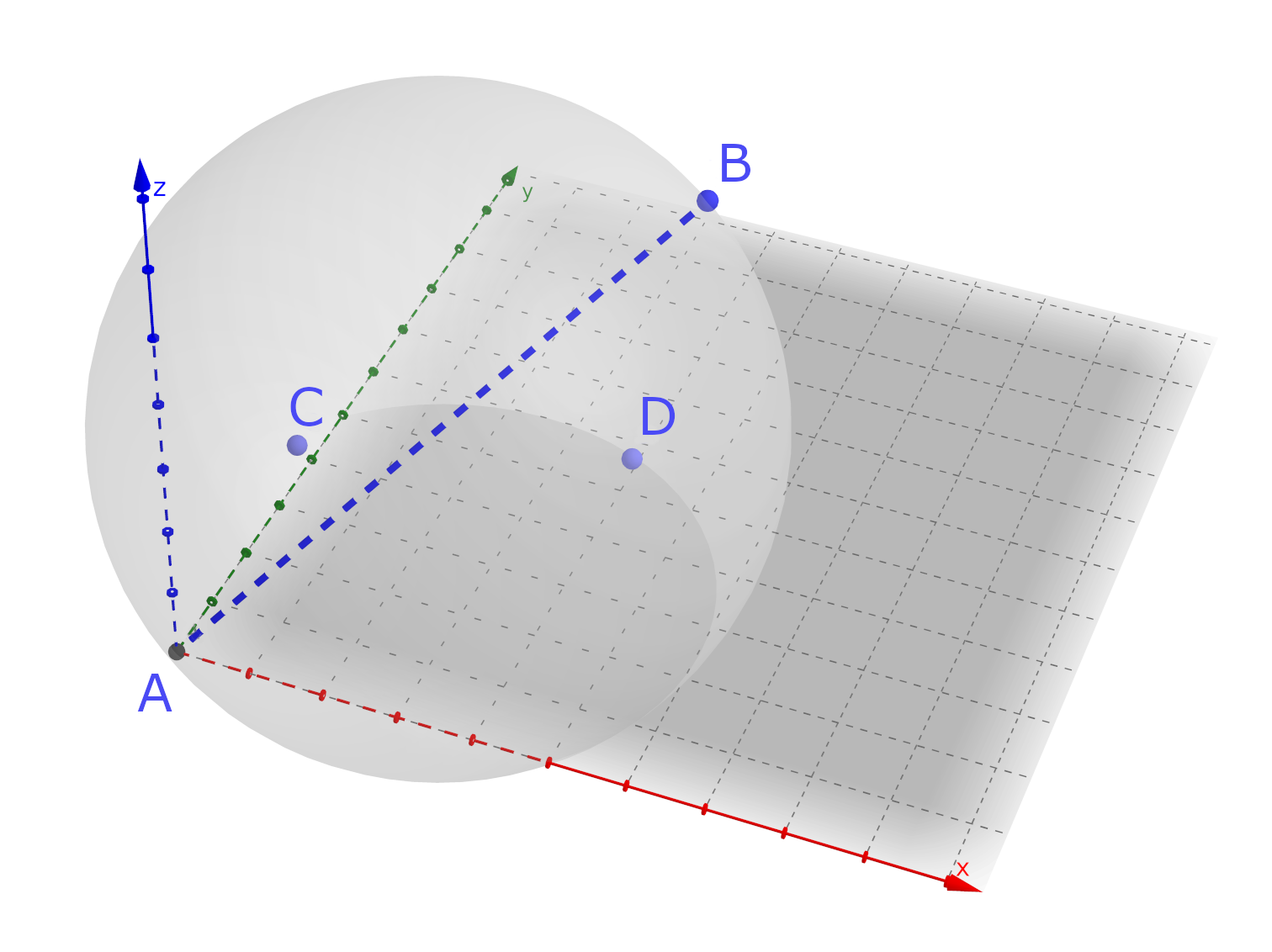}
  \includegraphics[width=0.3\linewidth]{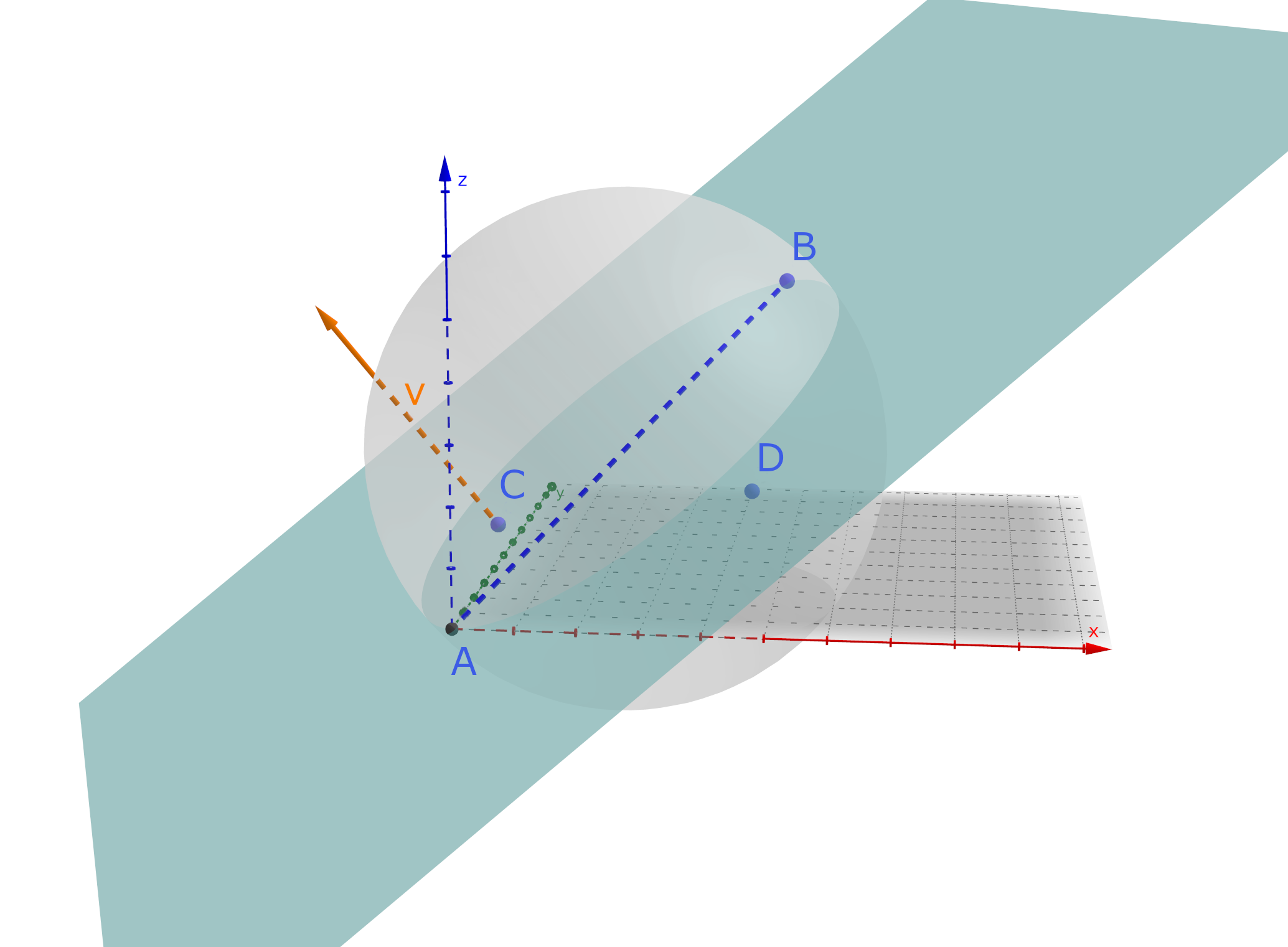}
  \includegraphics[width=0.3\linewidth]{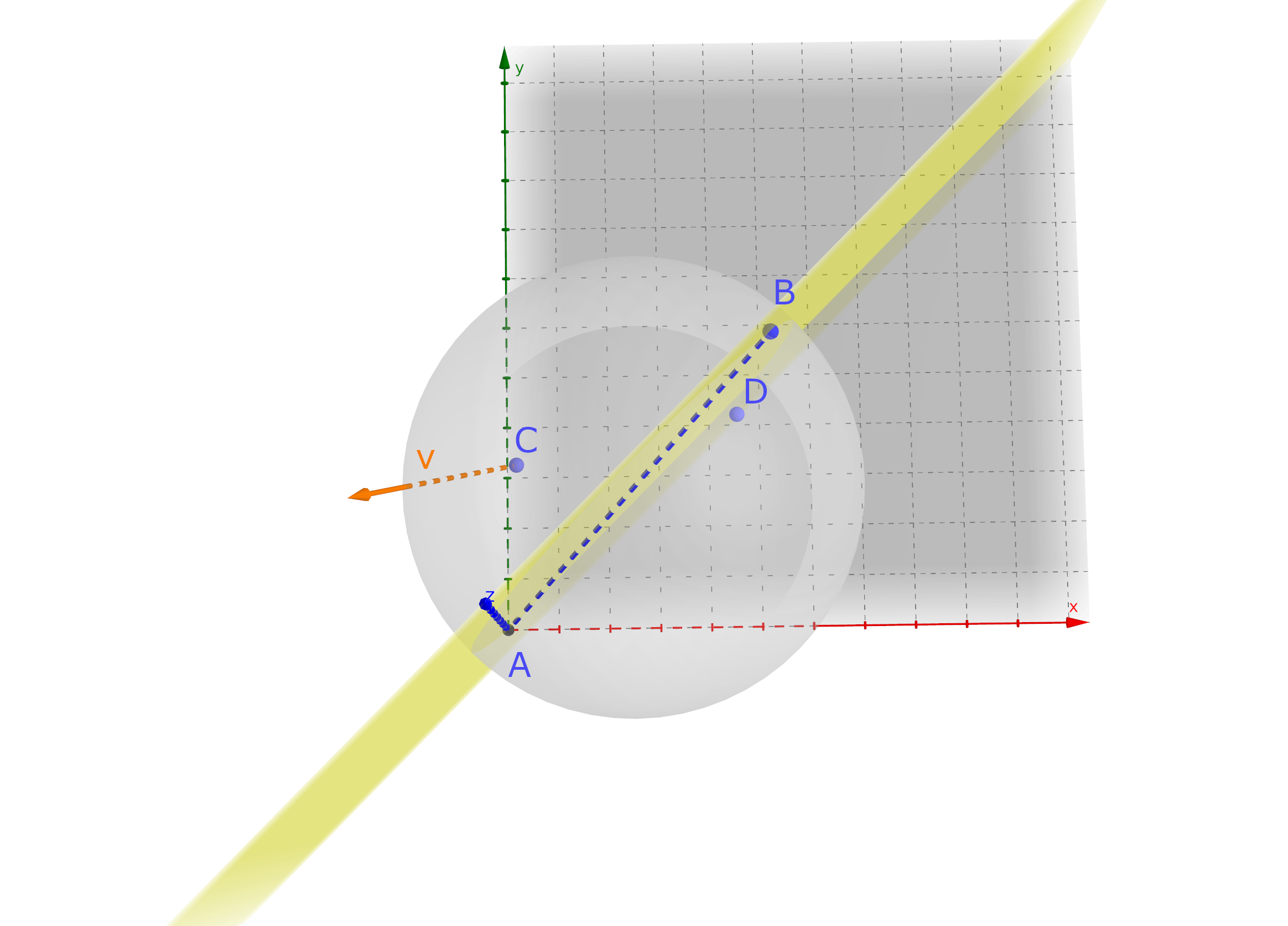}}
  \caption{Construction of a geometric descriptor for a quad of 3D ``stars''. (Left) Defining the coordinate system. (Middle) A side view of the scene. The plane $ABC$ is represented in blue, and vector $\overrightarrow{v}$ in orange. (Right) A top view of the scene. The yellow plane is the plane defined by the line $AB$ and the Z-axis.}
  \label{fig:3d-descriptor-ab-line}
\end{figure*}

Without loss of generality, to keep the notation simple, we will consider in the following text and figures that we have $k=4$ points (``stars'') named $A$, $B$, $C$ and $D$ in a constellation.
Stars $A$ and $B$ are chosen as the two most widely separated stars (Figure~\ref{fig:3d-descriptor-ab-line}), and define the coordinate system over which the remaining stars will be represented. 
$A$ is defined as the origin of such coordinate system, and $B$ as the point $(1,1,1)$, and we chose them such that $A$ is the closer one to the constellation centroid, in order to break ties.
The coordinates $(C_x, C_y, C_z, D_x, D_y, D_z)$ of the remaining points in this new coordinate system will constitute the hash code that describes the constellation. $C$ and $D$ are chosen such that $C_x < D_x$, in order to break symmetries.

However, the solution described so far is not unique, because there are infinite coordinate systems defined by the points $A=(0,0,0)$ and $B=(1,1,1)$ by rotation around the direction $(1,1,1)$.
Thus, we use a third point --- let it be star $C$ ---, non-collinear with $A$ and $B$, to define a plane $ABC$ with them. In order to reduce the computational error as much as possible, making the solution more robust to noise, and also to make the solution unique, we choose for star $C$ the one which is further away from the line defined by the points $A$ and $B$.
We then compute the normal of the plane $ABC$ --- let it be vector $\overrightarrow{v}$ ---, which is a vector that, in particular, is orthogonal to $\overrightarrow{AB}$  (Figure~\ref{fig:3d-descriptor-ab-line}, Middle). With these two vectors, we can uniquely define a coordinate system. Let us define the coordinate system such that the projection of $\overrightarrow{v}$ onto the $Z$-axis is maximum.

Depending on the position of the star $C$ relative to $A$ and $B$, and on how we compute the normal to the plane $ABC$, we may obtain one of two opposite solutions --- say, $\overrightarrow{v}$ or $-\overrightarrow{v}$. Therefore, to break this symmetry, we fix the coordinate system such that point $C = (C_x, C_y, C_z)$ appears ``to the left'' of the plane defined by the line $AB$ and the Z-axis, when viewed from a top-down view onto the XY plane, as seen in Figure~\ref{fig:3d-descriptor-ab-line} (Right), i.e. such that $C_x \leq C_y$. If this condition does not hold, we simply rotate the coordinate system by $\ang{180}$ around the direction of $\overrightarrow{AB}$, which is equivalent to swapping the normal $\overrightarrow{v}$ by $-\overrightarrow{v}$.

In summary, we compute a family of coordinate systems such that the points $A$ and $B$ are mapped to $A=(0,0,0)$ and $B=(1,1,1)$, and then we eliminate the infinite solutions that still exist by rotating the coordinate system around the direction $\overrightarrow{AB}$ such that the projection of $\overrightarrow{v}$ onto the $Z$-axis is maximum and $C_x \leq C_y$. This gives us a unique solution.

\smallskip

The problem of finding the rotation around the direction $\overrightarrow{AB}$ that maximizes the projection of $\overrightarrow{v}$ onto the Z$-$axis has a closed-form solution.
Given the vector $\overrightarrow{AB}$ defined by the stars $A$ and $B$ --- let's normalise it and call it $\overrightarrow{u} = \frac{\overrightarrow{AB}}{||\overrightarrow{AB}||} = \left( \frac{\sqrt{3}}{3}, \frac{\sqrt{3}}{3}, \frac{\sqrt{3}}{3} \right)$ --- and the vector $\overrightarrow{v}$, which is the normal to the plane $ABC$ and is also orthogonal to $\overrightarrow{u}$, we want to find the angle to rotate vector $\overrightarrow{v}$ around $\overrightarrow{u}$ that maximises the projection of $\overrightarrow{v}'$ --- the rotated version of $\overrightarrow{v}$ --- onto the $Z$-axis.

We may compute the rotated vector $\overrightarrow{v}'$ around a unit vector $\overrightarrow{u}$, given a rotation angle $\theta$, by using the Rodrigues' formula.
Moreover, since in our case we know that $\overrightarrow{u} \perp \overrightarrow{v}$, and therefore $\overrightarrow{u} \cdot \overrightarrow{v} = 0$, the formula simplifies to
\begin{equation}
\overrightarrow{v}' = \overrightarrow{v} cos(\theta) + (\overrightarrow{u} \times \overrightarrow{v})sin(\theta) \,.
\end{equation}

We are only interested in the third component
\begin{equation}
    v'_z = v_z cos(\theta) + \frac{\sqrt{3}}{3} (v_y - v_x) sin(\theta) \,,
\end{equation}
which we want to maximise. Computing the derivative and its zeros we get the closed-form solution for $\theta$
\begin{equation}
 \theta = atan \left( \frac{ \sqrt{3}(v_y - v_z) }{ 3 v_z} \right) \quad \text{,  with $cos(\theta) \neq 0$ .}
\end{equation}

\subsection{Matching fruits across videos}

Our complete pipeline to match fruits across videos is summarised in Figure~\ref{fig:teaser}.
First, we run a fruit tracking algorithm proposed in \cite{matos_tracking_2024_anonymized} on a video of a row of trees in an orchard to produce a sparse point cloud of 3D fruit centroids. Then, we define constellations of $k=5$ points by considering, for each fruit centroid, its $n=10$ nearest neighbours and aggregating them in combinations of $5$ points\footnote{Lower values for $k$ increase the number of false matches, i.e. different constellations which happen to have similar hash codes. On the other hand, higher values of $k$ make the method less robust to missing points, because all the $k$ points of a constellation must be seen in order to compute its hash and match it with other constellations. The parameter $n$ increases redundancy and robustness, at the expense of increased combinatorial complexity. In our tests, we found out that $k=5$ and $n=10$ work particularly well for this problem.}.
These combinations introduce redundancy and robustness to missing data: each fruit participates in more than one constellation, and if one fruit is missing in a later video, we still have other constellations that do not depend on it to match the remaining ones.

For each constellation, a STaR-i descriptor is computed and stored in a map, along with the corresponding fruit IDs. 

When we want to re-identify the fruits on a second video of the same trees, we run the fruit tracking algorithm on this second video to extract 3D fruit centroids and constellations, as before, and then we match these constellations with those stored in the previously built map by finding the most similar descriptor. Descriptors are compared by Euclidean distance.

Each matched constellation induces a one-to-one correspondence of its individual fruits between the two videos. However, since each fruit participates in several constellations, and some false matches may occur due to noise, we may end up with contradictory correspondences for the same fruit. We hence count the number of candidate correspondences between any pair of fruits in the two videos, store them in a very sparse matrix, and run the Hungarian algorithm to obtain a final and more robust correspondence between fruits.

This re-identification of fruits, although robust, may not be complete since, due to occlusions or noise, some fruits in the second video may not have participated in any matched constellation. However, a matched constellation also induces a 3D rigid transformation that can be used to align two 3D sparse point clouds of fruit centroids of both videos in a local neighbourhood of those fruits.
Therefore, we use \ac{RANSAC} to robustly compute this rigid transformation, given the matched fruits, and align the point clouds, after which we can re-identify the remaining fruits in the neighbourhood by their nearest neighbour.

\begin{figure}
    \centering
    \includegraphics[width=0.9\linewidth]{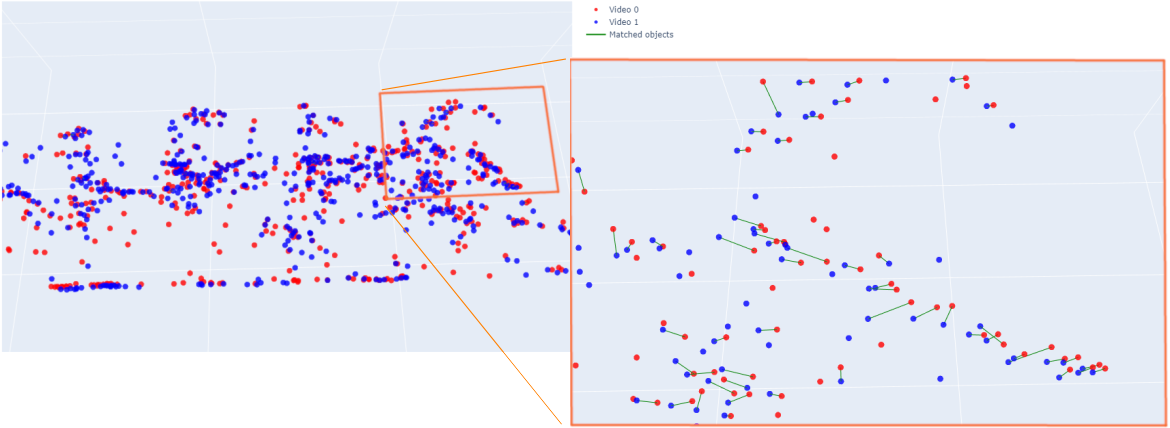}
    \caption{Registration of two complete semantic point clouds corresponding to two videos of the same trees. Matched fruits do not overlap perfectly.}
    \label{fig:point-cloud-registration}
\end{figure}

Note that a single rigid transformation that aligns the two complete point clouds of both videos may not exist, due to noise introduced by stereo and the object detector, as Figure~\ref{fig:point-cloud-registration} illustrates, hindering simpler approaches like point cloud registration through \ac{ICP}. Our method estimates several local transformations that explain the fruits that are visible on a given video frame at a time.

Finally, we apply a filtering technique proposed by Borges et al. \cite{borges_local_2020} on the 3D matched fruit centroids in order to eliminate possibly false correspondences, by modelling the Euclidean distances between them in a graph, by using those distances to rule out some edges, and by solving a \textit{maximum clique} problem. This eliminates geometrically inconsistent matches, and we found that it increased the Precision of our results without affecting their Recall.

\subsection{Application to robot navigation in an orchard}

The STaR-i descriptor induces a rigid 6 \ac{DoF} transformation between two matched constellations which, as described in the previous subsection, can be used with \ac{RANSAC} to find a more robust transformation between a camera (in the second video) and a map (built from the first video). 
Since, by lifting the problem to a semantic map of fruit centroids, we eliminated the dependency on visual features that may be affected by appearance or illumination changes, we can now localise a robot in an orchard map even under challenging conditions and despite changes in the trees over time.

It is worth noting that, although we used a calibrated stereo camera in this work, the method could work equally well with arbitrarily scaled 3D reconstructions derived from \ac{SfM} and a monocular camera, because the STaR-i descriptor is invariant to scale. Moreover, if the map was built in real-world scale, and the robot only has access to arbitrarily scaled point clouds from \ac{SfM}, we can still locate it in the real-world and even use the map to upgrade the \ac{SfM} reconstruction to a metric scale.

Finally, STaR-i descriptors can also be used to match frames in the same video and compute relative camera extrinsics (Figure~\ref{fig:camera_trajectory}), thus feeding a \ac{SLAM} system, for example.

\section{RESULTS}

\subsection{Descriptor robustness to occlusions and noise}

First, we want to assess the robustness of the constellation paradigm to occlusions, e.g. when some fruits that were previously seen and used to build constellations stored in a map cannot be detected in the current video frame, since this is a common issue with this kind of imagery \cite{matos_tracking_2024_anonymized}.

For this test, we used the video \textit{Galafab West} from a public dataset \cite{matos_tracking_2024_anonymized}. 
For each of the 259 video frames, we build an average of 5000 constellations of fruits, and then artificially remove some detections, to simulate occlusions, and we also add Gaussian noise with zero mean to the 3D positions of the fruit centroids.
Afterwards, we rebuild the constellations, try to match them with the ones initially built (without noise nor occlusions), and use those matches to compute the rigid transformation --- using Procrustes analysis with \ac{RANSAC} --- that transforms the perturbed 3D points into the original ones. 
In the end, we apply the estimated transformation to all the original points and measure the average Euclidean distance from each original 3D fruit centroid to its new position.
This experiment assesses the robustness of using the constellation paradigm to compute 3D rigid transformations under occlusions and/or noise.
All measurements are in meters. As a reference, the standard deviation range chosen for the noise covers the typical scale of an apple, and the whole scene spans approximately 10 meters in the longest dimension.

\begin{figure}[]
  \centerline{\includegraphics[width=0.8\linewidth]{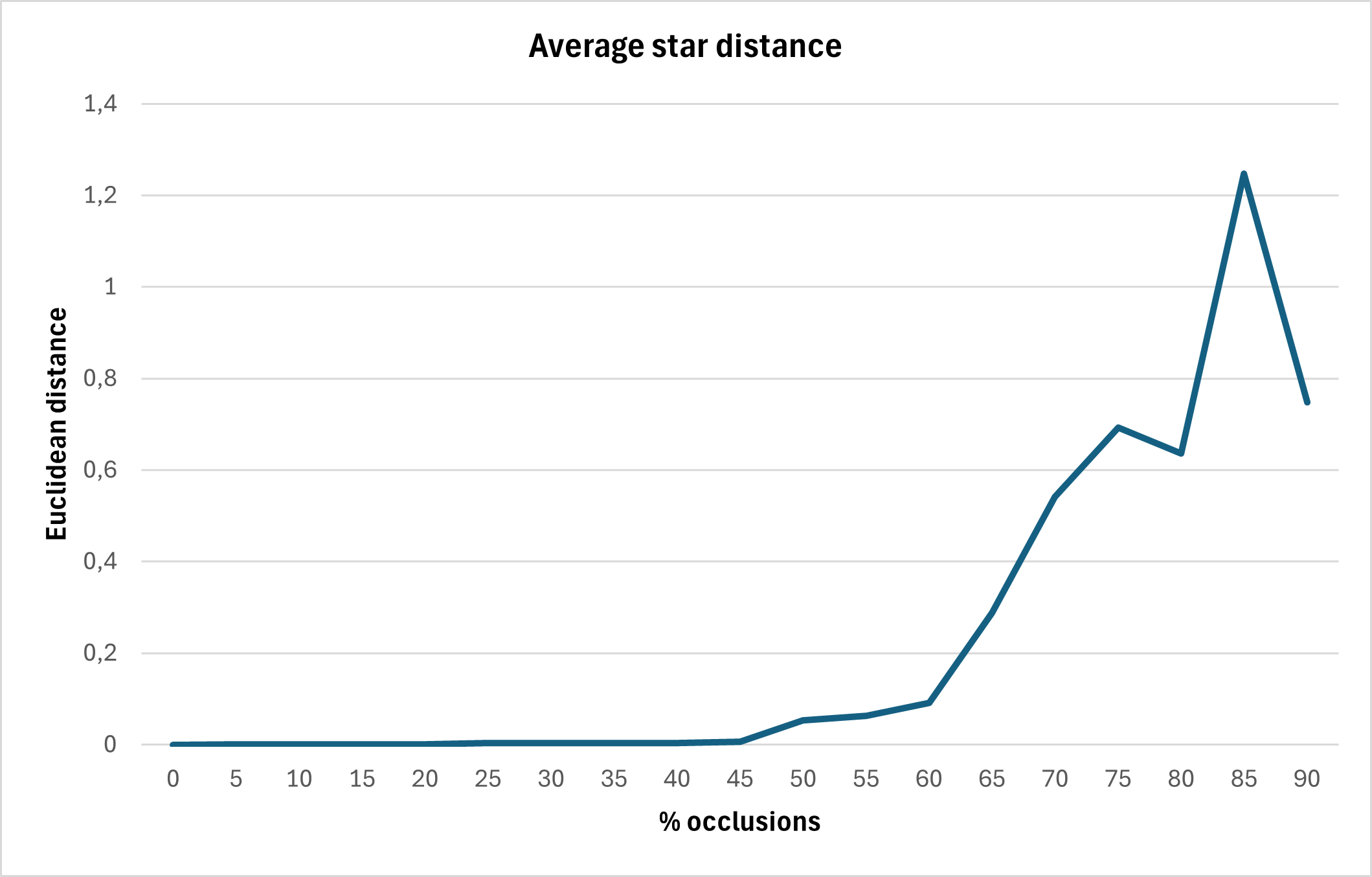}}
  \caption{Distance of the transformed 3D points to their original counterparts, as a function of the percentage of occlusions, without noise.}
  \label{fig:star-distance-occlusions-NN-10}
\end{figure}

\begin{figure}[]
  \centerline{\includegraphics[width=0.9\linewidth]{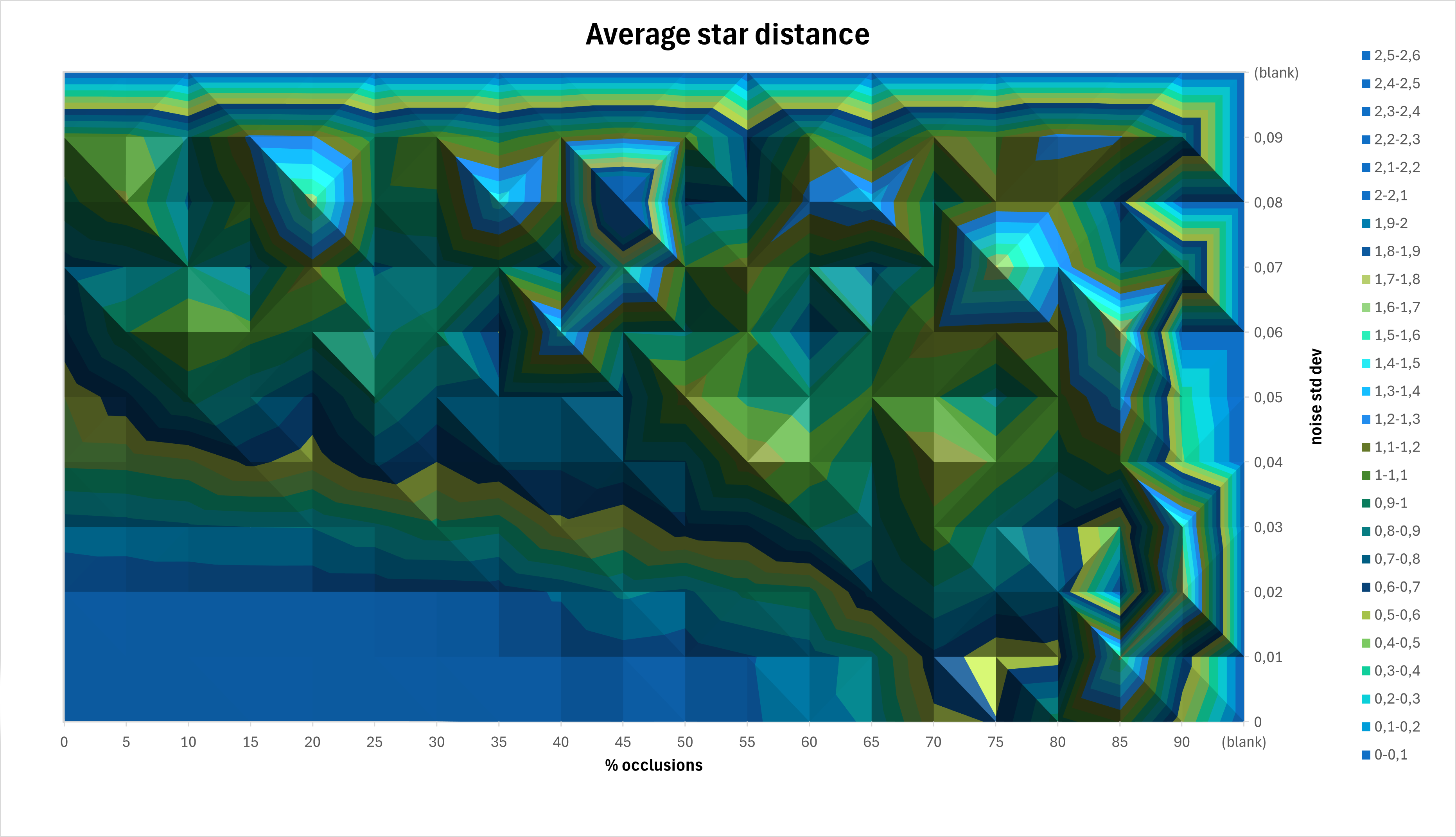}}
  \caption{Distance of the transformed 3D points to their original counterparts, as a function of the percentage of occlusions and standard deviation of the noise.}
  \label{fig:star-distance-occlusions-noise-NN-10}
\end{figure}

Figure~\ref{fig:star-distance-occlusions-NN-10} shows that, in the absence of noise, the estimated transformation is very stable regardless of the occlusion of fruits, up to $45\%$ of occluded fruits.
When we add noise to the experiment (Figure~\ref{fig:star-distance-occlusions-noise-NN-10}), we observe, as expected, that the results start to show perturbations earlier. Conversely, the robustness to noise diminishes as we increase the percentage of occlusions.

\subsection{Matching fruits across videos}

In order to evaluate our complete pipeline, we collected videos with a stereo camera over several weeks in an apple orchard owned by \ac{INIAV} \cite{iniav} in Alcobaça, Portugal, with varying light conditions, phenological states and camera trajectory/speed. We also generated some synthetic, photo-realistic videos, in order to assess the proposed method in a more controlled environment, without the noise introduced by stereo and \ac{SLAM}. The reader may watch all the datasets' videos in the accompanying video.

For each dataset, we run our algorithm on the first video (baseline), to produce a constellation map of fruits. Then, with this map as input, we run the algorithm over the same video (but backwards and skipping the first tree\footnote{This is to avoid a trivial solution where the camera starts exactly in the same pose as it was used to build the map.}), and over the remaining videos, to assess the robustness of the re-identification of fruits.
For each video, we either manually annotated the ID of each fruit matching those in the baseline video, or we got that data generated automatically in the case of synthetic videos, so we have ground truth matches. We can, hence, determine which correspondences are true positives, false positives\footnote{Fruits wrongly identified as one of the fruits of the baseline video.} and false negatives\footnote{Fruits which should have been identified as one of the fruits appearing in the baseline video, but which were not.}, and then compute Precision and Recall metrics.
We use a subset of each video covering 5 trees, 
and we only consider fruits that are visible at least on 5 frames of the video, to avoid possible false detections and noisy centroid estimations. Each result is an average of 10 runs with the exact same object detections as input.

The synthetic videos in Table~\ref{tab:results-synthetic} are 3D computer animations depicting 5 trees. The camera moves linearly, parallel to the line of trees, on the first two videos, and with a more erratic motion on the third video. The camera height and orientation changes between all the videos, allowing the fruits to be observed from different angles. The weather and light conditions also change in the third video.

\begin{table}
    \begin{tabular}{l|cc}
        Video & Precision & Recall \\
        \hline
         synthetic-apples-1 (baseline) (*) & 0.9366 & 0.9366  \\
         synthetic-apples-2 & 0.8712  & 0.8236 \\
         synthetic-apples-3 & 0.8668 & 0.8446  \\
    \end{tabular}
    \caption{Fruit matching results on synthetic scenarios. \textit{(*) The map was created from this video, and all results refer to fruit matches against this video.}}
    \label{tab:results-synthetic}
\end{table}

The baseline video in Table~\ref{tab:results-short-term} was captured with the camera mounted on a tractor. The second video was captured in the same manner, but with a faster motion of the tractor. The third video was captured by someone holding the camera while walking through the orchard, at a greater distance from the trees and in the opposite direction relative to the previous videos. The image is also darker, since the weather conditions changed. All videos were shot on the same day.

\begin{table}
    \begin{tabular}{l|cc}
        Video & Precision & Recall \\
        \hline
         tractor\_2024-08-28\_baseline (*) & 0.9405 & 0.9336  \\
         tractor\_2024-08-28\_faster & 0.8801  & 0.8570 \\
         walking\_2024-08-28\_darker\_far\_away & 0.4304 & 0.4419  \\
    \end{tabular}
    \caption{Fruit matching results on real-world, short-term scenarios.  \textit{(*) The map was created from this video, and all results refer to fruit matches against this video.}}
    \label{tab:results-short-term}
\end{table}

The results in both the synthetic scenario (Table~\ref{tab:results-synthetic}) and the real-world, short-term scenario (Table~\ref{tab:results-short-term}) show that the method can accurately identify the same fruits under similar settings (first two videos in each dataset), despite changes in motion direction, speed, and camera angle.
The third video in each dataset is more challenging, because the appearance of the scene changes more drastically due to weather, light, camera distance to trees, and camera angle (see the bottom images in Figure~\ref{fig:teaser}C, or the accompanying video). The drastic change in camera distance and angle make the centroid estimations (which are based on stereo) noisier, and the light conditions affect the object detector, leaving us with fewer valid constellations to match, which explains the poorer results.
Ideally, the constellation map would be used in a similar pose to how it was created, e.g. with the camera mounted on a tractor.

\begin{figure}
    \centering
    \includegraphics[width=0.80\linewidth]{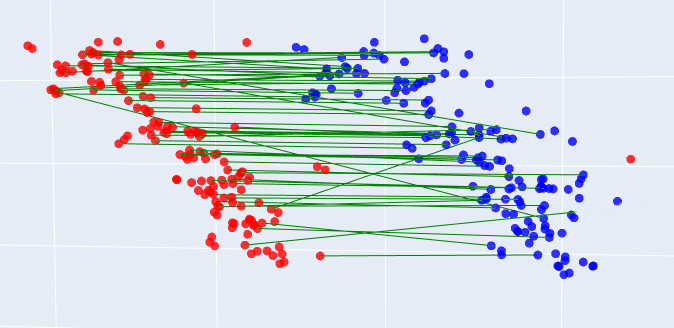}
    \caption{Fruit matches (green lines) between \textit{tractor\_2024-08-28\_baseline} (red point cloud) and \textit{walking\_2024-08-28} (blue point cloud), seen from above.}
    \label{fig:fruit-macthes-2024-08-28-baseline-walk}
\end{figure}

Figure~\ref{fig:fruit-macthes-2024-08-28-baseline-walk} shows that, despite the relatively low metrics of video \textit{walking\_2024-08-28}, approximately half of the fruits are matched and those matches are, in the majority of cases, either correct or at least matched with a very plausible fruit (e.g. a neighbour). These results are remarkable, given the image conditions, and are sufficient for a relatively good localisation in the orchard at tree level. Only a few matches, depicted by green lines that are not parallel to the others, are completely off-site, but these do not affect the pose estimated by \ac{RANSAC}.

\smallskip

We also evaluate the performance of our method while matching the same fruits captured on different dates. This poses new challenges due to the changes in appearance caused by different light and weather conditions, fruits' size and colour variation through the vegetative cycle.

The first video was shot on 9 August, by walking on foot. The second video was captured on 21 August with the camera in a tractor, turned upside down, and moving in opposite direction with respect to the first video. The third video was shot on 28 August, under settings similar to the second video.

\begin{table}
    \begin{tabular}{l|cc}
        Video & Precision & Recall \\
        \hline
         walking\_2024-08-09\_baseline (*) & 0.8973 & 0.8280  \\
         tractor\_2024-08-21\_inverse & 0.7092 & 0.7148 \\
         tractor\_2024-08-28\_inverse & 0.7775 & 0.7929  \\
    \end{tabular}
    \caption{Fruit matching results on real-world, long-term scenarios.  \textit{(*) The map was created from this video, and all results refer to fruit matches against this video.}}
    \label{tab:results-long-term}
\end{table}

Table~\ref{tab:results-long-term} shows, in comparison with Table~\ref{tab:results-short-term}, that the matching difficulty increases over time, as expected, since the plants grow, and consequently the scene changes in a non-rigid way, becoming less similar to the original map. Nevertheless, the method could still successfully match more than $70\%$ of the fruits (Figure~\ref{fig:matches_2024-0809-2024-0828_1}), which is sufficient to accurately localise a robot, for example, and determine a 6 \ac{DoF} camera pose with respect to the map.  

For the tasks requiring tracking fruits throughout a vegetative cycle, the constellation map should probably be redone or updated regularly, since after a few weeks of plant development we observe that the re-identification capability using the same map decreases.

The results in tables \ref{tab:results-short-term} and \ref{tab:results-long-term} also suggest that scenes captured with the camera in hand lead to less accurate maps and re-identifications than those shot with a tractor. Therefore, the process of mapping an orchard should preferably be done with a vehicle and possibly a stabilised camera.

\subsection{Localisation and navigation}

In this subsection, we show graphical evidence that our method can actually be used to locate the camera in 6 \ac{DoF} with respect to a previously built map, and also to estimate relative extrinsics between frames of the same video.

\begin{figure}
    \centering
    \includegraphics[width=0.60\linewidth]{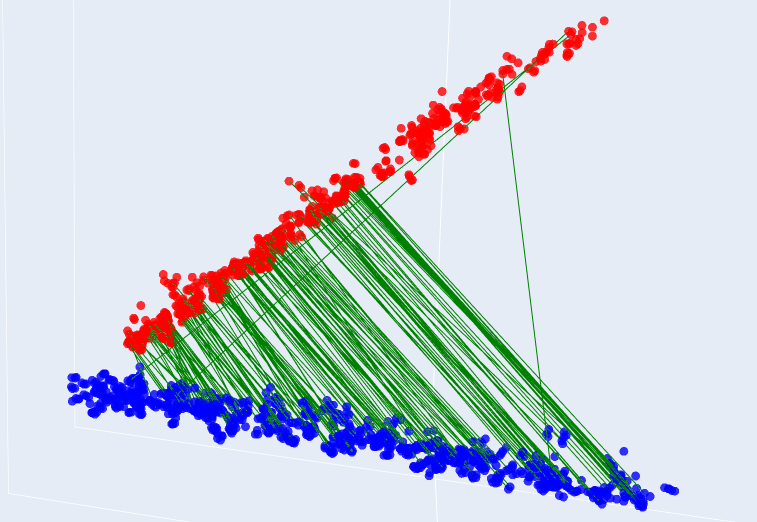}
    \caption{Fruit matches (green lines) between \textit{walking\_2024-08-09\_baseline} (red point cloud) and \textit{tractor\_2024-08-28\_inverse} (blue point cloud), seen from above.}
    \label{fig:matches_2024-0809-2024-0828_1}
\end{figure}

Figure~\ref{fig:matches_2024-0809-2024-0828_1} shows the matching between fruits in two point clouds, where the red one was used to build the constellation map and the blue one is being localised. 
It is a partial matching because the two videos cover different portions of the row of trees, having only some trees in common.
The point clouds are in arbitrary coordinate systems, as they are reconstructed, and were not registered to improve the visibility of the matching lines.
Here we may observe that the constellation paradigm correctly identifies the common fruits without any prior information about the camera starting pose nor location in the field.

\begin{figure}
    \centering
    \includegraphics[width=0.50\linewidth]{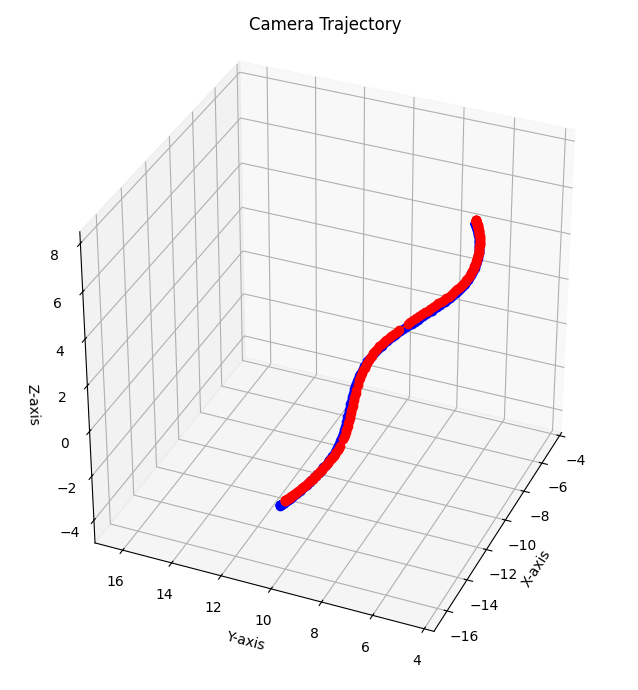}
    \caption{Ground truth (blue) and estimated (red) camera trajectory for the \textit{synthetic-apples-3} video.}
    \label{fig:camera_trajectory}
\end{figure}

Figure~\ref{fig:camera_trajectory} shows an estimated camera trajectory for the video \textit{synthetic-apples-3} along with its ground truth, which is almost coincident. The camera extrinsics were estimated using only the fruit matches provided by the constellation paradigm.
This result illustrates the applicability of this paradigm to \ac{SLAM} pipelines, for example, under challenging conditions where standard visual features like \ac{SIFT} may fail due to the the lack of distinctive and rigid features to track.

\section{CONCLUSIONS}

In this paper, we propose a new paradigm to address the problem of re-identification of fruits across videos of orchards under challenging conditions, which is a core necessity for other tasks in precision agriculture, like estimating fruit growth rates. We also contribute with STaR-i, a descriptor for very sparse 3D point clouds that is invariant to translation, rotation, and scale.

The short-term results presented here show that the constellation paradigm can successfully re-identify fruits even under scenarios where the visual appearance of objects change considerably due to non-rigidity of the scene, occlusions, and different illumination, camera pose and distance. Light conditions and camera pose, in particular, seem to be the most determinant causes of some worse results.

Preliminary results on long-term settings show that the method could also potentially re-identify fruits throughout the vegetative cycle if new maps are built regularly, although this concrete application needs more development and validation.

Moreover, we also show that the constellation paradigm can be used to estimate camera extrinsics and feed a \ac{SLAM} system for agricultural robots' navigation where other \ac{SLAM} systems, based on visual features, may fail due to challenging imagery. 

In future work, we intend to extend the constellation paradigm to tackle more of the long-term issues, making the map more robust to non-rigid deformations of the scene caused by fruits' growth over time.






\section*{ACKNOWLEDGMENT}

\iftoggle{anonimo}{
\textit{*Anonymized*}
}{
This work was partially funded by COMPETE2030 and FEDER programs under operation code COMPETE2030-FEDER-00630300 (project AI4OptiAgri).
This work was also supported by LARSyS funding (DOI: 10.54499/LA/P/0083/2020, 10.54499/UIDP/50009/2020, and 10.54499/UIDB/50009/2020) and 10.54499/2022.07849.CEECIND/CP1713/CT0001, through Fundação para a Ciência e a Tecnologia, and by the SmartRetail project [PRR - C645440011-00000062], through IAPMEI - Agência para a Competitividade e Inovação.
We thank to INIAV \cite{iniav}, for providing access to the orchards, for collecting images and data.
}



\AtNextBibliography{\small}
\printbibliography 

\begin{acronym}
\setlength{\parskip}{0ex}
\setlength{\itemsep}{1ex}
\itemsep=-5pt
\acro{AUC}[AUC]{Area Under the Curve}
\acro{ANN}[ANN]{Artificial Neural Network}
\acro{APE}[APE]{Absolute Percentage Error}
\acro{BA}[BA]{Bundle Adjustment}
\acro{CHT}[CHT]{Circular Hough Transform}
\acro{CNN}[CNN]{Convolutional Neural Network}
\acro{DBSCAN}[DBSCAN]{Density-Based Spatial Clustering of Applications with Noise}
\acro{DL}[DL]{Deep Learning}
\acro{DoF}[DoF]{Degrees of Freedom}
\acro{DSLR}[DSLR]{Digital Single-Lens Reflex}
\acro{EM}[EM]{Expectation-Maximisation}
\acro{FCN}[FCN]{Fully Convolutional Network}
\acro{FN}[FN]{False Negative}
\acro{FNN}[FNN]{Feedforward Neural Network}
\acro{FP}[FP]{False Positive}
\acro{FPR}[FPR]{False Positive Rate}
\acro{FPFH}[FPFH]{Fast Point Feature Histogram}
\acro{GA}[GA]{Genetic Algorithm}
\acro{GPU}[GPU]{Graphics Processing Unit}
\acro{GPS}[GPS]{Global Positioning System}
\acro{GMM}[GMM]{Gaussian Mixture Model}
\acro{GUI}[GUI]{Graphical User Interface}
\acro{HDRI}[HDRI]{High Dynamic Range Image}
\acro{HOTA}[HOTA]{Higher Order Tracking Accuracy}
\acro{ICP}[ICP]{Iterative Closest Point}
\acro{IMU}[IMU]{Inertial Measurement Unit}
\acro{INIAV}[INIAV]{Instituto Nacional de Investigação Agrária e Veterinária}
\acro{IoT}[IoT]{Internet-of-Things}
\acro{IoU}[IoU]{Intersection over Union}
\acro{ISR}[ISR]{Institute for Systems and Robotics}
\acro{KLT}[KLT]{Kanade-Lucas-Tomasi}
\acro{KL}[KL]{Kullback–Leibler}
\acro{KNN}[KNN]{K-Nearest Neighbour}
\acro{LiDAR}[LiDAR]{Light Detection And Ranging}
\acro{MAE}[MAE]{Mean Absolute Error}
\acro{mAP}[mAP]{Mean Average Precision}
\acro{MAPE}[MAPE]{Mean Absolute Percentage Error}
\acro{MHT}[MHT]{Multple Hypothesis Tracking}
\acro{MOT}[MOT]{Multiple Object Tracking}
\acro{MOTA}[MOTA]{Multiple Object Tracking Accuracy}
\acro{MVS}[MVS]{Multi-View Stereo}
\acro{MSAC}[MSAC]{M-estimator Sample Consensus}
\acro{NARF}[NARF]{Normal Aligned Radial Feature}
\acro{NBDL}[NBDL]{Neighbour-Binary Landmark Density}
\acro{NeRF}[NeRF]{Neural Radiance Field}
\acro{NIR}[NIR]{Near-Infrared}
\acro{NMS}[NMS]{Non-Maximum Suppression}
\acro{PCL}[PCL]{Point Cloud Library}
\acro{PDF}[PDF]{probability density function}
\acro{PnP}[PnP]{Perspective-n-Point}
\acro{RANSAC}[RANSAC]{RANdom SAmple Consensus}
\acro{RBF}[RBF]{Radial Basis Function}
\acro{RF}[RF]{Random Forest}
\acro{ROI}[ROI]{Region of Interest}
\acro{SfM}[SfM]{Structure-from-Motion}
\acro{SIFT}[SIFT]{Scale-Invariant Feature Transform}
\acro{SLAM}[SLAM]{Simultaneous Localisation and Mapping}
\acro{SOM}[SOM]{Self-Organising Map}
\acro{SURF}[SURF]{Speeded-Up Robust Features}
\acro{SVD}[SVD]{Singular Value Decomposition}
\acro{SVM}[SVM]{Support Vector Machine}
\acro{TN}[TN]{True Negative}
\acro{TP}[TP]{True Positive}
\acro{TPR}[TPR]{True Positive Rate}
\acro{UAV}[UAV]{Unmanned Autonomous Vehicle}
\acro{YOLO}[YOLO]{You Only Look Once}
\end{acronym} 

\end{document}